\RequirePackage[hyphens]{url}

\documentclass[11pt,a4paper]{article}
\usepackage[hyperref]{acl2021}
\usepackage{times}
\usepackage{latexsym}
\usepackage{nicefrac}

\PassOptionsToPackage{hyphens}{url}\usepackage{hyperref}

\usepackage{tikz}

\usepackage{enumitem}

\usepackage{microtype}

\usepackage{amsmath}
\usepackage{bbm}

\aclfinalcopy 

\title{Few-Shot Upsampling for Protest Size Detection}
  
\author{Andrew Halterman \\
  Massachusetts Institute of Technology \\
  \texttt{ahalt@mit.edu} \\\And
  Benjamin J. Radford \\
  UNC Charlotte \\
  \texttt{benjamin.radford@uncc.edu} \\}

\date{}

\begin{document}
\maketitle
\begin{abstract}
We propose a new task and dataset for a common problem in social science research: ``upsampling'' coarse document labels to fine-grained labels or spans. We pose the  problem in a question answering format, with the answers providing the fine-grained labels. We provide a benchmark dataset and baselines on a socially impactful task: identifying the exact crowd size at protests and demonstrations in the United States given only order-of-magnitude information about protest attendance, a very small sample of fine-grained examples, and English-language news text. We evaluate several baseline models, including zero-shot results from rule-based and question-answering models, few-shot models fine-tuned on a small set of documents, and weakly supervised models using a larger set of coarsely-labeled documents. We find that our rule-based model initially outperforms a zero-shot pre-trained transformer language model but that further fine-tuning on a very small subset of 25 examples substantially improves out-of-sample performance. We also demonstrate a method for fine-tuning the transformer span on only the coarse labels that performs similarly to our rule-based approach. This work will contribute to social scientists' ability to generate data to understand the causes and successes of collective action. 
\end{abstract}

\section{Introduction}

\begin{figure}[ht]
    \centering
    \includegraphics[width=\columnwidth]{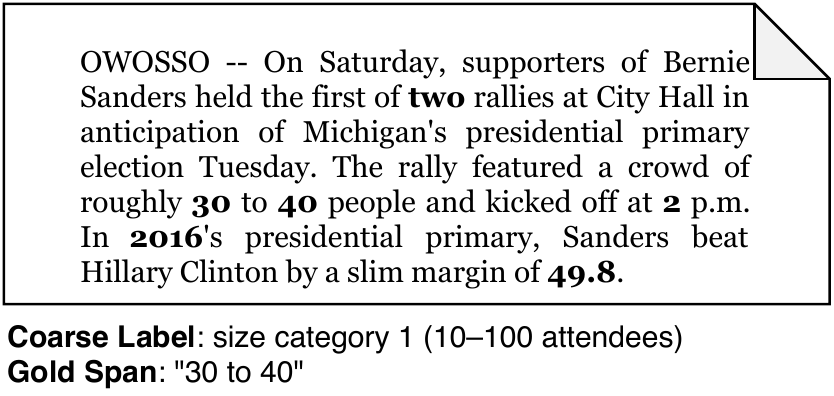}
    \caption{Documents in our corpus have ``coarse labels'' reporting the order of magnitude of the protest size and ``gold spans'' reporting the exact size of the protest. The frequency of number words (in bold) shows why this task is not trivial.}
    \label{fig:high_level}
\end{figure}

A common data collection task in social science is applying fine-grained labels to documents, including extracting specific passages from text. In many cases, social scientists already have many coarsely-labeled documents and a small number of hand-annotated documents. An automated technique for ``upsampling'' from coarse labels to more detailed information could help researchers produce better tailored datasets. However, this process does not fit the tools that applied researchers have access to: training a document classifier on coarse labels will not produce the fine-grained answers. Innovations in zero-shot and few-shot classifiers and information extraction (IE) techniques show promise, but new methods are required that can also draw on the existing coarse document annotations to improve fine-grained extraction.

We introduce a new task and dataset for improving information extraction systems' performance when given many coarsely-labeled documents and a small number of documents annotated with the spans of interest.\footnote{Replication archive available at \url{https://github.com/benradford/few-shot-upsampling-for-protest-size-detection}.} We draw on a dataset on dissent and collective action (hereafter, ``protests'') in the United States compiled by the \citet{ccc2020} (CCC) to construct our training and evaluation data. Protests are an important avenue for social change and of major interest for social science researchers. Current work suggests that attendance is a major factor in the success of a protest movement \citep{chenoweth:margherita:2019}, but good data on protest attendance is difficult to collect. CCC compiles structured data about protests from expert annotators using news reporting, including the exact text span from the article that describes the protest's size and the order of magnitude of the crowd size. An example is given in Figure~\ref{fig:high_level}. The task we propose is to locate the span within a document that reports the size of a protest, given a training set of documents labeled with the order of magnitude of the protest (``\textit{coarse labels}'') and a small number of document pieces (25) with exact span information (``\textit{gold spans}'').

Drawing on recent work in question answering, we repurpose existing models to generate fine-grained labels given a large set of coarsely-labeled documents and a small set of documents with fine-grained labels. We provide results from three baseline models, finding that a heuristic, rule-based system outperforms a zero-shot transformer-based question-answering (QA) model. Fine tuning on a small set (25) of gold spans substantially improves performance. We also introduce a new multitask model that reaches equivalent performance despite fine-tuning on \textit{no} gold spans.

\section{Task and Data}

For each protest in the CCC dataset, we collect the following data: the raw article text (scraped from the CCC-provided URLs), the exact string reporting the protest size, and a ``size category'' provided by CCC that reports the order of magnitude size of the crowd. The task is to predict the size text string, given plentiful training data with the size category and the gold spans for a small set of partial documents (25 paragraphs). The test set includes only the full article texts and order-of-magnitude information. To make the task tractable, we exclude protests that are coded from multiple documents and documents from which multiple protests are coded. From 48,736 total protests reported by CCC between January 21, 2017 and October 31, 2020, we eliminate multi-document/multi-protest reports and successfully scrape text for 11,005 protests. We eliminate documents where the CCC-reported size text is not located within the document, leaving 3,849 protests/documents. We split these data into four parts:

\begin{itemize}[noitemsep]
    \item \textbf{Coarse label training set}: text with coarse, order-of-magnitude labels \{0,1,2,3\} but no exact answer spans (2,694 full articles).
    \item \textbf{Gold span training set}: short texts with exact answer spans but no order-of-magnitude labels (25 paragraphs).
    \item \textbf{Validation set}: documents with order-of-magnitude labels and exact answer spans (200 full articles).
    \item \textbf{Test set}: documents with order-of-magnitude labels and exact answer spans (930 full articles).
\end{itemize}

The task is challenging because models are not evaluated on the largest portion of the data (coarse document labels) but rather on a fine-grained span prediction task for which only limited data is available. The task can thus be framed in several ways, depending on which parts of the data are used and in what ways:

\begin{itemize}[noitemsep]
    \item \textbf{Zero shot}: use an off-the-shelf model to detect protest sizes without any fine tuning on our data, either coarse or fine.
    \item \textbf{Few-shot on gold spans}: fine tune a baseline model on the small number of gold span labelled data. 
    \item \textbf{Coarse labels}: use a coarse-to-fine model to identify spans given only document-level labels.
    \item \textbf{Coarse labels + gold spans}: train a model using both coarse order-of-magnitude labels and limited fine-grained span data. 
\end{itemize}

\section{Related Work}

The task we propose relates to several strands of research. One framing is as a \textit{question-answering} task (QA), where the same question (``How many people protested?") is asked about each document. A large set of NLP tasks can be framed as question-answering models \citep{mccann2018natural} and QA models trained on language models can generalize to new domains with few or no labeled examples \citep{brown2020language, radford2019language}. QA models have also been successfully used when the training data is noisy \citep{lin2018denoising}. Given the flexibility of QA models and their strong performance in new domains, we use one as the base of our models. 

A different framing is as a \textit{``rationale"} problem for a document classifier. \citet{lei2016rationalizing} train a classifier on document-level labels and use attention weights to extract rationales for the classification. Our task differs from the canonical document classification task because a responsive model is evaluated on the extracted spans, not on the coarse label prediction task. 

\textit{Distant supervision} uses noisy labels, often applied automatically or with heuristic labels, to train systems \citep{ratner2017snorkel}. The classic example of distant supervision uses a database of relations to label binary relations in text \citep{mintz2009distant}. Weak supervision, more generally, uses labels that are noisy or coarse to train fine-grained models \citep{khetan2018learning, robinson2020strength}. Some work on ``noisy labels'' relates to our task, where labels are presented at a higher level of aggregation rather than with noise. 
\citet{nayak2020semi} propose a model that uses coarse, document-level sentiment labels to train a fine-grained, sentence-level sentiment classifier. Their task differs from ours in the nature of their labels: in moving from document-level to sentence-level labels, they predict labels of the same type (sentiment scores). In our task, we also change the labels themselves, from a crowd size order of magnitude to a token-level label of whether a word describes the exact protest size.

\newcommand{\tp}{\ensuremath{|\hat{\mathcal{T}_i} \cap \mathcal{T}_i|}}
\newcommand{\fp}{\ensuremath{|\hat{\mathcal{T}_i} \setminus \mathcal{T}_i|}}
\newcommand{\fn}{\ensuremath{|\mathcal{T}_i \setminus \hat{\mathcal{T}_i}|}}

\section{Modeling Strategy}

We first attempt the task using a rule-based model (the ``heuristic keyword model'') and an off-the-shelf zero-shot QA system. 
We then introduce a multi-task neural network model based on a pre-trained transformer language model. We fine-tune and evaluate this model on the  coarse labels and gold spans, as well as on noisy labels we generate through a rule-based procedure.

The two standard performance metrics for question answering tasks are exact match and F$_1$ \citep{rajpurkar2018know}. We compute exact match as the sum of exact matches (predicted spans exactly matched in the set of correct target spans) divided by the total number of documents. We compute F$_1$ per document based on token-level precision and recall, then average across documents. 

\subsection{Heuristic Keyword Model}

Our heuristic model is a rule-based system that uses keyword matching and dependency parses to return a single number-containing phrase from the article. We first locate all number-containing phrases (digits or number words) in the text with regular expressions. Using a rule-based system, we convert these number phrases to a numeric form (e.g. ``several dozen" $\rightarrow$ 36) and then compare the phrase's numerical value to the protest's reported order of magnitude. If the phrase does not match the order of magnitude, we eliminate it from our candidate list. To further reduce the candidate list, we look for number phrases that occur within the same sentence as a set of keywords such as ``crowd'', ``gathered'', or ``protesters''.\footnote{The complete list is ``protesters'', ``demonstrators'', ``gathered'', ``crowd", ``rallied'', ``attended'', ``picketed'', ``protest''.} If multiple sentences have keyword matches, we return the first one. The CCC data's size spans include modifiers alongside the raw numerical values (e.g. ``\textit{about} 20", ``\textit{more than} 50"). We use dependency parse information generated by spaCy to extract the wider span.\footnote{Specifically, (1) for each sentence matching a keyword (2) identify the word in the sentence that is a number word or numeric, and (3) also include child nodes that had the following labels: adjectival modifier, modifier of quantifier, compound, adverbial modifier. We used spaCy version 2.3.2 with the \texttt{en\_core\_web\_lg} model to perform the dependency parsing and  sentence segmentation.}

\subsection{Zero-Shot QA Model}

\begin{figure}
    \centering
    \includegraphics[width=1\linewidth]{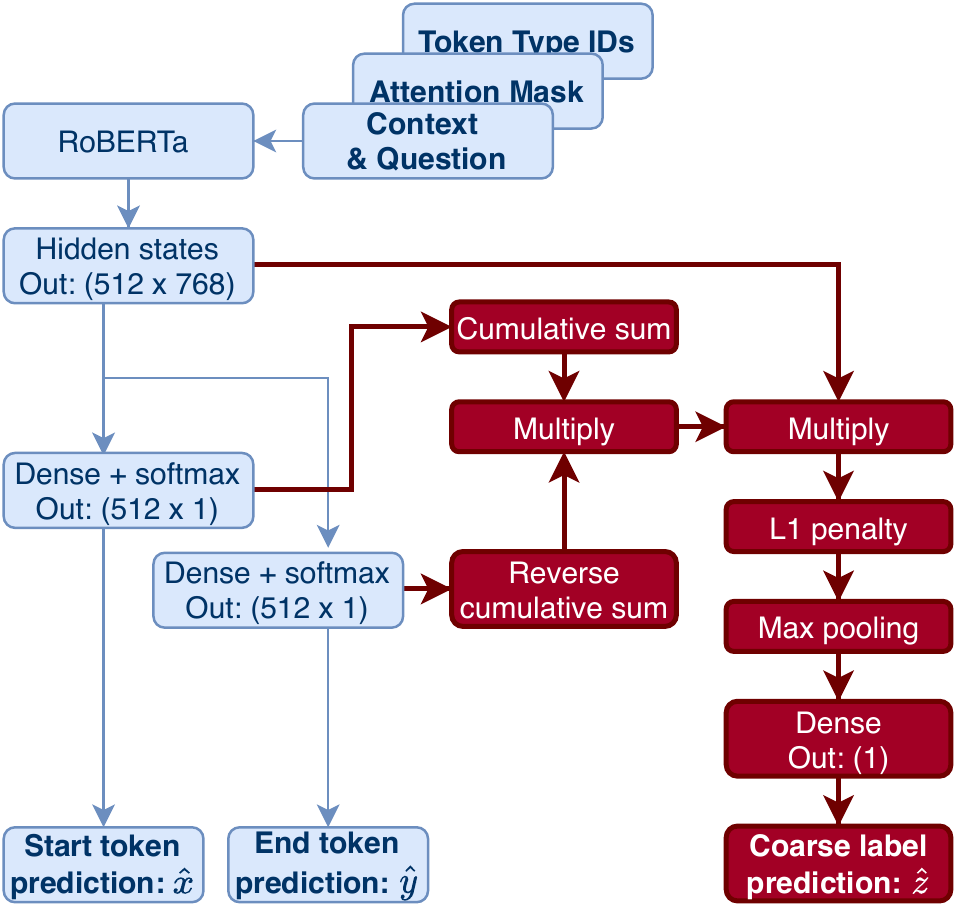}
    \caption{Multitask model architecture: standard RoBERTa QA (left) and attention mask-based regression for coarse label prediction (right).}
    \label{fig:architecture}
\end{figure}

We begin with a pre-trained RoBERTa model \citep{liu:etal:2019} that we subsequently fine-tune  for question answering using the Stanford Question Answering Dataset (SQuAD) 2.0  as described in Appendix~\ref{appendix:roberta} \citep{rajpurkar:etal:2018}.\footnote{We use \texttt{roberta-base} from HuggingFace \citep{wolf2020transformers}.}  The QA model architecture is depicted on the left side of Figure~\ref{fig:architecture}. Because we do not tune this model on our dataset, we consider its predictions to be zero-shot.

\begin{table}[ht!]
    \centering
   \begin{tabular}{@{\extracolsep{0pt}}l r r} 
    \hline
       \textbf{Model} & {\textbf{Exact}} & {\textbf{F$_1$}} \\
        \hline
        {Heuristic rules} &  0.54 &  0.61 \\
        {RoBERTa QA}  && \\
        {\hspace{0pt}zero-shot} & 0.17 & 0.27\\
        {\hspace{0pt}+ gold spans}      & 0.67 & 0.65 \\
        {\hspace{0pt}\emph{+ coarse labels}}   & 0.48 & 0.54 \\
        {\hspace{0pt}\emph{+ coarse labels + heuristic spans}}  & 0.66 & 0.63 \\
         \hline
    \end{tabular}
    \caption{Exact match (``Exact'') and F$_1$ performance on test set data. All RoBERTa QA and multitask models are fine-tuned on SQuAD 2.0. Multitask models italicized. Full results given in Appendix~\ref{appendix:results}.}
    \label{tab:results}
\end{table}

\subsection{Fine-tuned QA Model}\label{sec:fine}

To use the coarse labels, we add an additional objective to the QA model that is trained to predict the crowd size order of magnitude. The model first predicts the start and end token vectors for a given context-question pair. We compute the cumulative sum (over tokens) of the predicted start token vector and the reverse cumulative sum for the predicted end token vector. 
The resulting vectors are element-wise multiplied to produce an attention mask with high values in the range of tokens between the predicted start and end tokens. 
We apply an L1 penalty to this mask to ensure the attention focuses on a small number of 
tokens. The attention mask is then element-wise multiplied with the token hidden states produced by RoBERTa. Global max pooling and a single linear regression layer  applied to these attended-to hidden states predict the coarse label (as shown in the right side of Figure~\ref{fig:architecture}).

The loss function for the multitask model, an unweighted combination of cross-entropy loss and mean squared error, is \(
- \sum_{i=1}^{n} \left\{ x_i \text{log}(\hat{x}_i) 
     + y_i \text{log}(\hat{y}_i) \right\}
     + (\hat{z}-z)^2\), 
where $x_i \in \{0,1\}$ indicates whether token $i$ is the start of an answer span, $y_i \in \{0,1\}$ indicates whether token $i$ is the end of an answer span, $z$ is the document's coarse label, and $n$ is the number of tokens (512, here). The model can be fit to data including any combination of these three targets.


\section{Results}

Results on the test set are given in Table~\ref{tab:results}. RoBERTa QA refers to RoBERTa fine-tuned on SQuAD 2.0. With only fine-tuning on SQuAD 2.0, the model scores 17\% exact match accuracy and 27\% F$_1$. 
On their own, the heuristic-derived spans outperform zero-shot RoBERTa QA.
``+ Heuristic spans'' indicates that the given model was fine-tuned on the spans identified by the heuristic model. 

\begin{figure}
    \centering
    \includegraphics[width=1\linewidth]{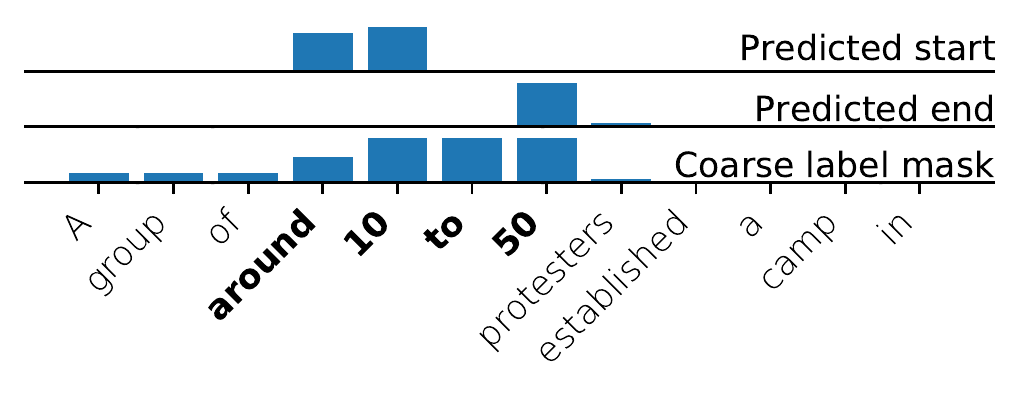}
    \caption{Example target span from document excerpt with predicted start tokens (top), predicted end tokens (middle), and attention mask (bottom). Results from model \textit{c} in Table~\ref{tab:results}. Actual span in bold.}
    \label{fig:example}
\end{figure}

Fine-tuning the multitask model on the coarse labels alone results in a 180\% increase in exact match accuracy and 100\% increase in F-score. An example prediction made by the multitask coarse labels model is shown in Figure~\ref{fig:example}.\footnote{The model just misses an exact match by omitting ``around'' from the predicted span.} However, the highest scores are achieved by fine-tuning the RoBERTa QA model on just the 25 gold spans: 67\% exact match  accuracy and 65\% F-score.

The greatest performance by a multitask model without any gold spans is achieved by the model fine-tuned on both the coarse labels and the heuristic spans: 66\% exact match  and 63\% F$_1$, just below the top performing model with access to the gold spans. We interpret the success of this model and the coarse labels model over the base RoBERTa QA model as evidence that our attention masking strategy was successful at upsampling from coarse document-level labels to specific token-level spans. 

\section{Discussion and Conclusion}

Social scientists often find themselves with coarsely-labeled text data for which upsampling may provide valuable additional information.  We anticipate applications in extracting fine-grained policy proposals from party manifestos with document-level annotations \citep{manifestos2017}, the specific armed actors engaged in civil war violence from documents labeled with ``rebel'' or ``government'' \citep{lyall2010coethnics}, or the specific phrases in news text that lead to their censorship \citep{king2013censorship}. We also see applications in upsampling ranges of causalities from NGO reports or Wikipedia articles to the exact sizes, upsampling years to more specific dates, or using rounded numbers from financial disclosures or government reports as coarse supervision for extracting the exact amount from text.

Improvements in zero- and low-shot models should encourage applied researchers to explore computational approaches to text analysis even when training data is scarce, noisy, or coarse---common challenges that are often perceived as intractable.
At the same time, NLP researchers should continue to improve models that can learn to extract fine-grained information given coarse training data. Multitask QA models show promise in doing so, but future work can further integrate work from the weak/distant supervision literature, including modeling the noisiness of the labels.



\section*{Acknowledgments}

We thank Jay Ulfelder of the Crowd Counting Consortium for helpful discussions.
Halterman gratefully acknowledges the support of an NSF Graduate Research Fellowship.

\section*{Impact Statement}

Studies of protests have the potential for serious ethical concerns. Some tasks, such as identifying or de-anonymizing the participants in a protest could produce major harms. Our application, identifying the number of attendees at a protest, has less potential for harm. Our  collection of information on the size of protests will generally accord with the desires of protesters. Social scientists have long seen protests as an important tool for social movements to overcome collective action problems: by making support for a position visible in the streets, a protest assures potential supporters of the protest that their opinions are held by others and that the group could potentially achieve its ends with more support \citep{kuran1989sparks, petersen2001resistance, tarrow2011power}. Providing better information on the size of protests furthers the signalling and information-disseminating objectives of the protesters themselves. While we might not agree with the causes of all protesters in the United States, we believe that on-balance, our work benefits those with less power more than it does those with greater power, who can likely already collect the information they seek manually. 

The data that we draw on was collected by the Crowd Counting Consortium, which relies on volunteers and paid research assistants to collect the data. Their protocol was reviewed by the University of Denver IRB and deemed exempt because they do not collect personally identifiable information and use only public data.\footnote{\url{https://sites.google.com/view/crowdcountingconsortium/faqs}}

A second consideration in our work involves the role of copyrighted news text in our project. Our method uses copyrighted news text that we scraped from the web. While scraping websites is legal in the United States,\footnote{https://www.eff.org/deeplinks/2019/09/victory-ruling-hiq-v-linkedin-protects-scraping-public-data} redistributing copyrighted text is more difficult to justify and depends on how the use fits into the fair use doctrine. Balancing copyright holders' rights with public and educational benefit is at the core of the fair use doctrine.\footnote{\url{https://www.copyright.gov/fair-use/more-info.html}} Our attempt to balance the harms to copyright holders and the harms to broader public and scientific benefit is to publish a URL list and scraper so that our corpus can be re-created by future researchers. Additionally, in cases where a researcher is attempting to replicate our work for educational purposes, we will make our scraped corpus available for the narrow purpose of replicating our work.

%

\bibliographystyle{acl_natbib}
\bibliography{acl2021}

\begin{thebibliography}{24}
\expandafter\ifx\csname natexlab\endcsname\relax\def\natexlab#1{#1}\fi

\bibitem[{Abadi et~al.(2015)Abadi, Agarwal, Barham, Brevdo, Chen, Citro,
  Corrado, Davis, Dean, Devin, Ghemawat, Goodfellow, Harp, Irving, Isard, Jia,
  Jozefowicz, Kaiser, Kudlur, Levenberg, Man\'{e}, Monga, Moore, Murray, Olah,
  Schuster, Shlens, Steiner, Sutskever, Talwar, Tucker, Vanhoucke, Vasudevan,
  Vi\'{e}gas, Vinyals, Warden, Wattenberg, Wicke, Yu, and
  Zheng}]{tensorflow:2015}
Mart\'{\i}n Abadi, Ashish Agarwal, Paul Barham, Eugene Brevdo, Zhifeng Chen,
  Craig Citro, Greg~S. Corrado, Andy Davis, Jeffrey Dean, Matthieu Devin,
  Sanjay Ghemawat, Ian Goodfellow, Andrew Harp, Geoffrey Irving, Michael Isard,
  Yangqing Jia, Rafal Jozefowicz, Lukasz Kaiser, Manjunath Kudlur, Josh
  Levenberg, Dan Man\'{e}, Rajat Monga, Sherry Moore, Derek Murray, Chris Olah,
  Mike Schuster, Jonathon Shlens, Benoit Steiner, Ilya Sutskever, Kunal Talwar,
  Paul Tucker, Vincent Vanhoucke, Vijay Vasudevan, Fernanda Vi\'{e}gas, Oriol
  Vinyals, Pete Warden, Martin Wattenberg, Martin Wicke, Yuan Yu, and Xiaoqiang
  Zheng. 2015.
\newblock \href {http://tensorflow.org/} {{TensorFlow}: Large-scale machine
  learning on heterogeneous systems}.
\newblock Software available from tensorflow.org.

\bibitem[{Brown et~al.(2020)Brown, Mann, Ryder, Subbiah, Kaplan, Dhariwal,
  Neelakantan, Shyam, Sastry, Askell et~al.}]{brown2020language}
Tom~B Brown, Benjamin Mann, Nick Ryder, Melanie Subbiah, Jared Kaplan, Prafulla
  Dhariwal, Arvind Neelakantan, Pranav Shyam, Girish Sastry, Amanda Askell,
  et~al. 2020.
\newblock Language models are few-shot learners.
\newblock \emph{arXiv preprint arXiv:2005.14165}.

\bibitem[{Chenoweth and Margherita(2019)}]{chenoweth:margherita:2019}
Erica Chenoweth and Belgioioso Margherita. 2019.
\newblock \href
  {https://librarylink.uncc.edu/login?url=https://www-proquest-com.librarylink.uncc.edu/scholarly-journals/physics-dissent-effects-movement-momentum/docview/2369964731/se-2?accountid=14605}
  {The physics of dissent and the effects of movement momentum}.
\newblock \emph{Nature Human Behaviour}, 3(10):1088--1095.
\newblock Copyright - 2019© The Author(s), under exclusive licence to Springer
  Nature Limited 2019; Last updated - 2020-03-25.

\bibitem[{{Crowd Counting Consortium}(2020)}]{ccc2020}
{Crowd Counting Consortium}. 2020.
\newblock \href {https://sites.google.com/view/crowdcountingconsortium/home}
  {[link]}.

\bibitem[{Khetan et~al.(2018)Khetan, Lipton, and
  Anandkumar}]{khetan2018learning}
Ashish Khetan, Zachary~C Lipton, and Animashree Anandkumar. 2018.
\newblock Learning from noisy singly-labeled data.
\newblock In \emph{International Conference on Learning Representations}.

\bibitem[{King et~al.(2013)King, Pan, and Roberts}]{king2013censorship}
Gary King, Jennifer Pan, and Margaret~E Roberts. 2013.
\newblock How censorship in china allows government criticism but silences
  collective expression.
\newblock \emph{American Political Science Review}, 107(2):326--343.

\bibitem[{Kuran(1989)}]{kuran1989sparks}
Timur Kuran. 1989.
\newblock Sparks and prairie fires: A theory of unanticipated political
  revolution.
\newblock \emph{Public choice}, 61(1):41--74.

\bibitem[{Lehmann et~al.(2017)Lehmann, Matthie{\ss}, Merz, Regel, and
  Werner}]{manifestos2017}
Pola Lehmann, Theres Matthie{\ss}, Nicolas Merz, Sven Regel, and Annika Werner.
  2017.
\newblock Manifesto corpus. version: 2017b.
\newblock Technical report, Berlin: WZB Berlin Social Science Center.

\bibitem[{Lei et~al.(2016)Lei, Barzilay, and Jaakkola}]{lei2016rationalizing}
Tao Lei, Regina Barzilay, and Tommi Jaakkola. 2016.
\newblock Rationalizing neural predictions.
\newblock In \emph{Proceedings of the 2016 Conference on Empirical Methods in
  Natural Language Processing}, pages 107--117.

\bibitem[{Lin et~al.(2018)Lin, Ji, Liu, and Sun}]{lin2018denoising}
Yankai Lin, Haozhe Ji, Zhiyuan Liu, and Maosong Sun. 2018.
\newblock Denoising distantly supervised open-domain question answering.
\newblock In \emph{Proceedings of the 56th Annual Meeting of the Association
  for Computational Linguistics (Volume 1: Long Papers)}, pages 1736--1745.

\bibitem[{Liu et~al.(2019)Liu, Ott, Goyal, Du, Joshi, Chen, Levy, Lewis,
  Zettlemoyer, and Stoyanov}]{liu:etal:2019}
Yinhan Liu, Myle Ott, Naman Goyal, Jingfei Du, Mandar Joshi, Danqi Chen, Omer
  Levy, Mike Lewis, Luke Zettlemoyer, and Veselin Stoyanov. 2019.
\newblock \href {http://arxiv.org/abs/1907.11692} {Roberta: {A} robustly
  optimized {BERT} pretraining approach}.
\newblock \emph{CoRR}, abs/1907.11692.

\bibitem[{Lyall(2010)}]{lyall2010coethnics}
Jason Lyall. 2010.
\newblock Are coethnics more effective counterinsurgents? evidence from the
  second chechen war.
\newblock \emph{American Political Science Review}, 104(01):1--20.

\bibitem[{McCann et~al.(2018)McCann, Keskar, Xiong, and
  Socher}]{mccann2018natural}
Bryan McCann, Nitish~Shirish Keskar, Caiming Xiong, and Richard Socher. 2018.
\newblock The natural language decathlon: Multitask learning as question
  answering.
\newblock \emph{arXiv preprint arXiv:1806.08730}.

\bibitem[{Mintz et~al.(2009)Mintz, Bills, Snow, and
  Jurafsky}]{mintz2009distant}
Mike Mintz, Steven Bills, Rion Snow, and Dan Jurafsky. 2009.
\newblock Distant supervision for relation extraction without labeled data.
\newblock In \emph{Proceedings of the Joint Conference of the 47th Annual
  Meeting of the ACL and the 4th International Joint Conference on Natural
  Language Processing of the AFNLP: Volume 2-Volume 2}, pages 1003--1011.
  Association for Computational Linguistics.

\bibitem[{Nandan(2020)}]{nandan:2020}
Apoorv Nandan. 2020.
\newblock \href {https://keras.io/examples/nlp/text_extraction_with_bert/}
  {Text extraction with bert}.

\bibitem[{Nayak et~al.(2020)Nayak, Ghosh, Jia, Mithafi, and
  Kumar}]{nayak2020semi}
Guruprasad Nayak, Rahul Ghosh, Xiaowei Jia, Varun Mithafi, and Vipin Kumar.
  2020.
\newblock Semi-supervised classification using attention-based regularization
  on coarse-resolution data.
\newblock In \emph{Proceedings of the 2020 SIAM International Conference on
  Data Mining}, pages 253--261. SIAM.

\bibitem[{Petersen(2001)}]{petersen2001resistance}
Roger~D Petersen. 2001.
\newblock \emph{Resistance and rebellion: lessons from Eastern Europe}.
\newblock Cambridge University Press.

\bibitem[{Radford et~al.(2019)Radford, Wu, Child, Luan, Amodei, and
  Sutskever}]{radford2019language}
Alec Radford, Jeffrey Wu, Rewon Child, David Luan, Dario Amodei, and Ilya
  Sutskever. 2019.
\newblock Language models are unsupervised multitask learners.
\newblock \emph{OpenAI blog}, 1(8):9.

\bibitem[{Rajpurkar et~al.(2018{\natexlab{a}})Rajpurkar, Jia, and
  Liang}]{rajpurkar2018know}
Pranav Rajpurkar, Robin Jia, and Percy Liang. 2018{\natexlab{a}}.
\newblock Know what you don't know: Unanswerable questions for squad.
\newblock \emph{arXiv preprint arXiv:1806.03822}.

\bibitem[{Rajpurkar et~al.(2018{\natexlab{b}})Rajpurkar, Jia, and
  Liang}]{rajpurkar:etal:2018}
Pranav Rajpurkar, Robin Jia, and Percy Liang. 2018{\natexlab{b}}.
\newblock \href {http://arxiv.org/abs/1806.03822} {Know what you don't know:
  Unanswerable questions for squad}.

\bibitem[{Ratner et~al.(2017)Ratner, Bach, Ehrenberg, Fries, Wu, and
  R{\'e}}]{ratner2017snorkel}
Alexander Ratner, Stephen~H Bach, Henry Ehrenberg, Jason Fries, Sen Wu, and
  Christopher R{\'e}. 2017.
\newblock Snorkel: Rapid training data creation with weak supervision.
\newblock In \emph{Proceedings of the VLDB Endowment. International Conference
  on Very Large Data Bases}, volume~11, page 269. NIH Public Access.

\bibitem[{Robinson et~al.(2020)Robinson, Jegelka, and
  Sra}]{robinson2020strength}
Joshua Robinson, Stefanie Jegelka, and Suvrit Sra. 2020.
\newblock Strength from weakness: Fast learning using weak supervision.
\newblock In \emph{International Conference on Machine Learning}, pages
  8127--8136. PMLR.

\bibitem[{Tarrow(2011)}]{tarrow2011power}
Sidney~G Tarrow. 2011.
\newblock \emph{Power in movement: Social movements and contentious politics}.
\newblock Cambridge University Press.

\bibitem[{Wolf et~al.(2020)Wolf, Chaumond, Debut, Sanh, Delangue, Moi, Cistac,
  Funtowicz, Davison, Shleifer et~al.}]{wolf2020transformers}
Thomas Wolf, Julien Chaumond, Lysandre Debut, Victor Sanh, Clement Delangue,
  Anthony Moi, Pierric Cistac, Morgan Funtowicz, Joe Davison, Sam Shleifer,
  et~al. 2020.
\newblock Transformers: State-of-the-art natural language processing.
\newblock In \emph{Proceedings of the 2020 Conference on Empirical Methods in
  Natural Language Processing: System Demonstrations}, pages 38--45.

\end{thebibliography}

\newpage
\appendix

\begin{table*}[ht!]
    \centering
    \begin{tabular}{@{\extracolsep{6pt}}c l c c @{}}
    \hline
        & & Parameters & Training Time \\ 
        \hline
        (a) & {Heuristic rules} & -- & -- \\
        (b) & {RoBERTa + SQuAD 2.0 (zero-shot)} &   1.25M & 200 min \\
        (c) & {\hspace{12pt}  + coarse labels}     &  1.25M & + 20 min\\
        (d) & {\hspace{12pt}  + heuristic spans}     &  1.25M&  + 20 min\\
        (e) & {\hspace{12pt}  + coarse labels + heuristic spans}  & 1.25M&  + 20 min \\
        (f) & {\hspace{12pt}  + \textit{gold spans}}            &  1.25M &  + 20 min\\
        (g) & {\hspace{12pt}  + \textit{gold spans} + coarse labels}   &  1.25M&  + 20 min\\
        (h) & {\hspace{12pt}  + \textit{gold spans} + heuristic spans}  &  1.25M&  + 20 min\\
        (i) & {\hspace{12pt}  + \textit{gold spans} + coarse labels + heuristic spans}   &  1.25M&  + 20 min\\
         \hline
    \end{tabular}
    \caption{Model size in parameters. Training time (approximate) on 2$\times$ RTX 2080 Ti GPUs. ``+ 20 min'' indicates the model takes an additional 20 minutes to fine-tune after the initial fine-tuning on SQuAD 2.0. These estimates may be high due to our validation set performance evaluation between batches.}
    \label{tab:parameters}
\end{table*}

\begin{table*}[ht!]
    \centering
    \begin{tabular}{@{\extracolsep{6pt}}c l c c c c @{}}
    \hline
        & &  \multicolumn{2}{c}{Test set} & \multicolumn{2}{c}{Validation set} \\ \cline{3-4} \cline{5-6}
       && {Exact Match} & {F$_1$} & {Exact Match} & {F$_1$} \\
        \hline
        (a) & {Heuristic rules} &  0.54 &  0.61 & & \\
        (b) & {RoBERTa + SQuAD 2.0 (zero-shot)} &  0.17 & 0.27 & 0.19 & 0.27 \\
        (c) & {\hspace{12pt}  + coarse labels}     & 0.48 & 0.54 & 0.54 & 0.58 \\
        (d) & {\hspace{12pt}  + heuristic spans}     & 0.51 & 0.50 & 0.56 & 0.51 \\
        (e) & {\hspace{12pt}  + coarse labels + heuristic spans}  & 0.66 & 0.63 & 0.72 & 0.66 \\
        (f) & {\hspace{12pt}  + \textit{gold spans}}            & 0.67 & 0.65 & 0.71 & 0.68 \\
        (g) & {\hspace{12pt}  + \textit{gold spans} + coarse labels}   & 0.67 & 0.65 & 0.68 & 0.66 \\
        (h) & {\hspace{12pt}  + \textit{gold spans} + heuristic spans}  & 0.62 & 0.61 & 0.66 & 0.62 \\
        (i) & {\hspace{12pt}  + \textit{gold spans} + coarse labels + heuristic spans}   & 0.65 & 0.64 & 0.72 & 0.67 \\
         \hline
    \end{tabular}
    \caption{Exact match and token-level F$_1$ performance by each model on test and validation set data.}
    \label{tab:results_full}
\end{table*}

\section{Fine-tuning RoBERTa on SQuAD 2.0}\label{appendix:roberta}

\subsection{SQuAD 2.0 Fine-Tuning}

In order to facilitate extensions to the standard QA model, we perform the fine-tuning of RoBERTa on SQuAD 2.0 ourselves \citep{tensorflow:2015}. We fine-tune on the SQuAD 2.0 training set for three epochs using the settings recommended by \citet{nandan:2020}. We use a batch size of 12 due to memory limitations. We use the Adam optimizer with a learning rate of $5e-5$. Our model achieves 0.78 and 0.74 exact match  on the training and evaluation sets, respectively. We  use this model only as a basis for subsequent fine-tuning and therefore do not attempt to match state-of-the-art performance on the SQuAD 2.0 evaluation set. The model is trained on two RTX 2080 Ti GPUs. Model size and training time details are provided in Table~\ref{tab:parameters}.

We allow the QA model to identify impossible-to-answer questions by predicting the sequence start token (``\textless s\textgreater'') as both the answer span start and end token.

To fit within the RoBERTa base model's 512 token limit, we pre-process all text inputs via a shingling procedure. We limit contexts to 450 tokens thereby allowing questions of up to 62 tokens in length. We then pad to a uniform 512 tokens. When contexts exceed 450 tokens, we use a sliding window of 450 tokens that we step through the context 225 tokens at a time. We guarantee all samples generated from large contexts contain precisely 450 tokens by adjusting the first and last window positions such that they do not extend before or after the first or last context token, respectively. We aggregate predictions across shingles by assuming one predicted span per document and selecting the predicted span from the shingle for which $\text{max}_{i\in[1,\dots,512]}(\hat{x_i}) + \text{max}_{i\in[1,\dots,512]}(\hat{y_i})$ is the greatest.

\subsection{Task-Specific Fine-Tuning}

The selection of learning rate for these models, 5e-6 (exactly one order of magnitude lower than the default used for SQuAD fine-tuning), was due to our sensitivity to overfitting on the very small set of span examples. All models were trained for 150 batches, each batch comprising 12 samples chosen from the training datasets with replacement. When multiple datasets are used to train the same model, batches alternate between them. We selected the number of batches for training by observing exact match accuracy on the validation set over a range of iteration steps from 1 to 400 and selecting the earliest batch iteration at which validation set accuracy appeared to plateau.

\section{Results}\label{appendix:results}

The full set of fine-tuning data combinations is given in Table~\ref{tab:results_full}. All models \textit{c} through \textit{i} are trained using the same hyperparameters and strategy (Adam optimizer, 5e-6 learning rate, and 150 batches of size 12 examples each).

\end{document}